\documentclass[conference]{IEEEtran}
\IEEEoverridecommandlockouts
\usepackage{cite}
\usepackage{amsmath,amssymb,amsfonts}
\usepackage{algorithmic}
\usepackage{graphicx}
\usepackage{textcomp}
\usepackage{multirow}
\usepackage{xcolor}
\usepackage{pifont}
\usepackage{orcidlink}
\usepackage{url}

\newcommand{\red}[1]{\textcolor{red}{#1}}

\definecolor{green}{RGB}{0,200,0}
\definecolor{blue}{RGB}{0,120,255}

\begin{document}

\title{\vspace*{0.25in}MQE: Unleashing the Power of Interaction with \underline{M}ulti-agent \underline{Q}uadruped \underline{E}nvironment
}

\author{\IEEEauthorblockN{Ziyan Xiong$^{*\orcidlink{0009-0001-3319-4679}}$}
\and
\IEEEauthorblockN{Bo Chen}
\and
\IEEEauthorblockN{Shiyu Huang$^{\dag\orcidlink{0000-0003-0500-0141}}$}
\and
\IEEEauthorblockN{Wei-Wei Tu}
\and
\IEEEauthorblockN{Zhaofeng He}
\and
\IEEEauthorblockN{Yang Gao}
\thanks{Ziyan Xiong and Yang Gao are with the Institute for Interdisciplinary Information Sciences, Tsinghua University, Beijing, China (e-mail: xiongzy20@mails.tsinghua.edu.cn, gaoyangiiis@mail.tsinghua.edu.cn).}
\thanks{Bo Chen and Zhaofeng He are with AI Institute, Beijing University of Posts and Telecommunications, Beijing, China (e-mail: \{chenbo, zhaofenghe\}@bupt.edu.cn).}
\thanks{Shiyu Huang is with Zhipu AI, Beijing 100084, China (e-mail: shiyu.huang@aminer.cn).}
\thanks{Wei-Wei Tu is with 4Paradigm Inc., Beijing 100084, China (e-mail: tuweiwei@4paradigm.com).}
\thanks{Yang Gao is with Shanghai Artificial Intelligence Laboratory and Shanghai Qi Zhi Institute, Shanghai, China.}
\thanks{$^*$Work was done during the intern at 4Paradigm Inc..}
\thanks{$^\dag$Work was done while at 4Paradigm Inc..}
}

\maketitle
\begin{abstract}
The advent of deep reinforcement learning (DRL) has significantly advanced the field of robotics, particularly in the control and coordination of quadruped robots. However, the complexity of real-world tasks often necessitates the deployment of multi-robot systems capable of sophisticated interaction and collaboration. To address this need, we introduce the Multi-agent Quadruped Environment (MQE), a novel platform designed to facilitate the development and evaluation of multi-agent reinforcement learning (MARL) algorithms in realistic and dynamic scenarios. MQE emphasizes complex interactions between robots and objects, hierarchical policy structures, and challenging evaluation scenarios that reflect real-world applications. We present a series of collaborative and competitive tasks within MQE, ranging from simple coordination to complex adversarial interactions, and benchmark state-of-the-art MARL algorithms. Our findings indicate that hierarchical reinforcement learning can simplify task learning, but also highlight the need for advanced algorithms capable of handling the intricate dynamics of multi-agent interactions. MQE serves as a stepping stone towards bridging the gap between simulation and practical deployment, offering a rich environment for future research in multi-agent systems and robot learning. For open-sourced code and more details of MQE, please refer to \url{https://ziyanx02.github.io/multiagent-quadruped-environment/}.
\end{abstract}


\section{Introduction}

In recent years, the field of robotics has witnessed remarkable advancements in the locomotion control of quadruped robots. These strides have been largely propelled by the utilization of deep reinforcement learning (DRL) in simulation environments~\cite{makoviychuk2021isaac, rudin2022learning}, enabling quadrupeds to perform a wide array of tasks with unprecedented agility and efficiency, from trespassing~\cite{lee2020learning, miki2022learning, gai2024continual} rugged terrains to carrying out manipulative tasks~\cite{ji2022hierarchical, ji2023dribblebot, fu2022learning}, the capabilities of current quadrupeds represent a significant leap forward in robotic autonomy and versatility. 

Despite these advancements, the intricacy and variety of real-world tasks often surpass the capability of single-robot systems. These systems encounter fundamental constraints in performing tasks in situations such as large-scale environmental monitoring~\cite{nayyar2018smart, nguyen2023utilizing} and intricate logistical challenges within warehousing environments~\cite{pandian2019artificial, li2022supervised, lee2021mobile}, highlighting the critical need for the development and deployment of multi-robot systems. Such scenarios demand a level of coordination, collaboration, and perception that single-robot systems, no matter how complex they are, cannot achieve on their own.

Addressing the need for coordinated multi-robot systems, the research community has done much research on multi-agent reinforcement learning in different multi-agent environments, including the Multi-Agent Particle Environment (MPE)~\cite{lowe2017multi,liu2021sa}, Google Research Football~\cite{googlefootball,lin2023tizero}, and the StarCraft Multi-Agent Challenge (SMAC)~\cite{samvelyan2019starcraft,kim2022starcraft}. While these environments have significantly contributed to the advancement of MARL by providing platforms for developing and evaluating algorithms, we still lack tangible intermediaries to apply multi-agent policies to real-world tasks. The simplified or even frictional interactions in these environments often fall short of replicating the complex interference and dynamics present in real-world scenarios, such as imperfect action execution caused by direct or indirect interactions among real robots. These issues highlight a gap between the current research efforts and practical real-world applications.

\begin{figure}[t]
\centerline{\includegraphics[width=0.48\textwidth]{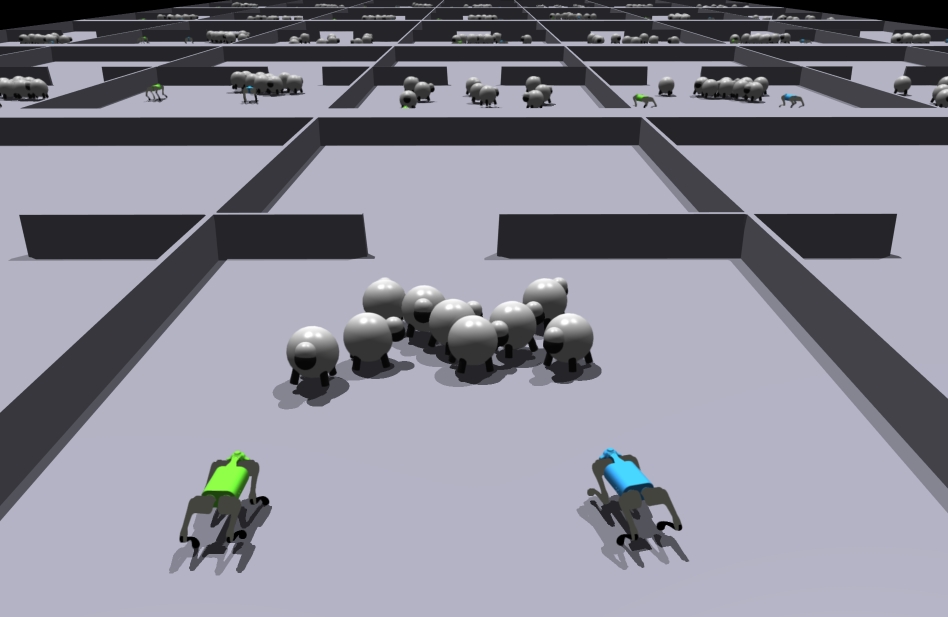}}
\caption{Agents learning to herd sheep in hundreds of parallel environments.}
\label{Overview}
\end{figure}

In response to these limitations, we propose the Multi-agent Quadruped Environment to bridge this gap, emphasizing 1) complex and dynamic interactions between objects and robots, 2) a hierarchical structure of policies that transforms complex control problems into simple multi-agent tasks, and 3) realistic and challenging evaluation scenarios of MARL focusing on either assessing the viability of simplifying complex multi-robot tasks through the hierarchical structure of policies or exploring robust locomotion policies amidst the unpredictable interactions among robots.

To explore the capabilities of our environment and the effectiveness of current MARL methodologies, we experiment with multi-agent reinforcement learning algorithms, comparing those with a hierarchical structure of policies intended to streamline the learning against those without such a structure. Our findings reveal that even advanced algorithms struggle to solve the more difficult tasks presented in our environment. This underscores not only the complexity and realism of the challenges we introduce but also the pressing need for our environment as a tool for developing more sophisticated and capable RL algorithms. As inspiration for future works, we also showcase several promising research directions in this environment, i.e. hierarchical RL and RL enhanced by self-play strategies.

\section{Related Works}

\subsection{Reinforcement Learning for Quadruped Control} 
The exploration of learning-based locomotion control has significantly advanced the capabilities of quadruped robots, enabling them to perform dynamic tasks such as robust walking on complex terrains~\cite{lee2020learning, agarwal2023legged, tsounis2020deepgait}, high-speed running~\cite{jin2022high,peng2020learning, hwangbo2019learning, margolis2022rapid, hutter2016anymal}, executing parkour abilities~\cite{cheng2023extreme}, climbing on obstacles~\cite{agarwal2023legged, fankhauser2018robust, lee2006quadruped, austin2019navigation}, and playing football~\cite{martins2021rsoccer, jidi_football, ji2022hierarchical}. These achievements highlight the importance of simulation environments~\cite{heess2017emergence, haarnoja2018learning} in facilitating the development of these complex locomotion strategies. Various simulators such as Gazebo, PyBullet, IsaacGym, and MATLAB have been employed, each offering different levels of fidelity, ease of use, and computational efficiency. Notably, IsaacGym, with its GPU-accelerated architecture, stands out for its enhanced sample efficiency, which is crucial for the rapid training of robots. 

Despite these advancements, the focus has largely been on single-agent scenarios. However, as tasks become more complex and demand higher levels of interaction, the need for multi-agent simulation environments becomes evident. The emergence of multi-robot simulation environments in other domains, such as drones~\cite{shah2018airsim, song2021flightmare, koch2019reinforcement} and dexterous manipulators~\cite{chen2023bi, todorov2012mujoco} highlights the necessity for similar advancements in quadruped robotics. Addressing this, our work introduces a novel multi-agent quadruped environment, aiming to bridge this gap and facilitate the exploration of cooperative and competitive dynamics among robots, thus expanding the scope of quadruped research into multi-robot operations and opening new avenues for the application of these technologies.

\subsection{Multi-agent Reinforcement Learning} 
Multi-agent reinforcement learning has increasingly received much research interest, propelled by its potential to tackle complex tasks through the collaborative and competitive interactions of multiple agents. The field distinguishes between two principal categories of MARL algorithms: value-based and policy-based. Value-based algorithms, including QMIX~\cite{rashid2018qmix}, VDN\cite{sunehag2017value}, and QPLEX\cite{wang2020qplex}, focus on estimating the value of actions in given states to determine the best course of action. On the other hand, policy-based algorithms like MADDPG\cite{lowe2017multi}, MAPPO\cite{yu2022surprising}, and MAT\cite{wen2022multi} directly learn a policy that maps states to actions, potentially offering more flexibility in continuous action spaces.

The development of MARL algorithms is established on the environments used for training and testing. Current MARL environments, such as MPE and SMAC, often present simplified, idealized scenarios, lacking complicated dynamics, i.e. in robotics simulation, to evaluate the capability of MARL algorithms. However, existing multi-robot environments primarily assess MARL algorithms through navigation tasks~\cite{wen2022cl,yang2023learning, surmann2020deep}, which significantly differ from the simplified settings commonly employed in MARL research. This discrepancy highlights the necessity for integrating both kinds of environments.

\section{Multi-agent Quadruped Environment}

The Multi-agent Quadruped Environment is developed atop the widely recognized reinforcement learning framework for legged robots, \emph{legged\_gym}, which operates on NVIDIA's Isaac Gym. An overview of MQE can be found in Fig. \ref{Overview}, accompanied by a comparative analysis of MQE against existing legged robot simulations in Table \ref{Comparison}. In addition to the wide range of quadruped models available through \emph{legged\_gym}, MQE introduces several notable features:

\begin{table}[h]
\caption{Comparison between Quadruped Environments}
\begin{center}
\begin{tabular}{cccc}
\hline
 & Multi-agent & Hierarchical policy & Simulator \\
\hline
\hline
champ & \textcolor{green}{\ding{51}} & \textcolor{green}{\ding{51}} & Gazebo \\
Rex-gym & \textcolor{green}{\ding{51}} & \textcolor{red}{\ding{55}} & PyBullet \\
Genloco & \textcolor{red}{\ding{55}} & \textcolor{red}{\ding{55}} & PyBullet \\
\emph{legged\_gym} & \textcolor{red}{\ding{55}} & \textcolor{green}{\ding{51}} & IsaacGym \\
\hline
MQE (ours) & \textcolor{green}{\ding{51}} & \textcolor{green}{\ding{51}} & IsaacGym \\
\hline
\end{tabular}
\label{Comparison}
\end{center}
\end{table}

\paragraph{Multi-robot interaction} Our environment is designed to seamlessly incorporate actuated robots, stationary and manipulable objects, and non-player characters (NPCs) which are objects that adhere to predetermined policies within each parallel-simulated environment. The inclusion of multiple robots and interactive objects intensifies the complexity of learning as each robot, or agent must navigate interactions with others, necessitating collaboration or evasion to mitigate the impacts of those interactions.

\paragraph{Modular terrain registration} Similar to many other legged robot environments, the simulated ground is segmented into parallel tracks to facilitate domain randomization of terrain or to implement curriculum learning strategies. Notably, our environment further divides each track into blocks, simplifying the creation and adaptation of terrain for various tasks. These blocks can be efficiently repurposed and reconfigured to meet diverse requirements.

\begin{figure*}[h]
\centerline{\includegraphics[width=\textwidth]{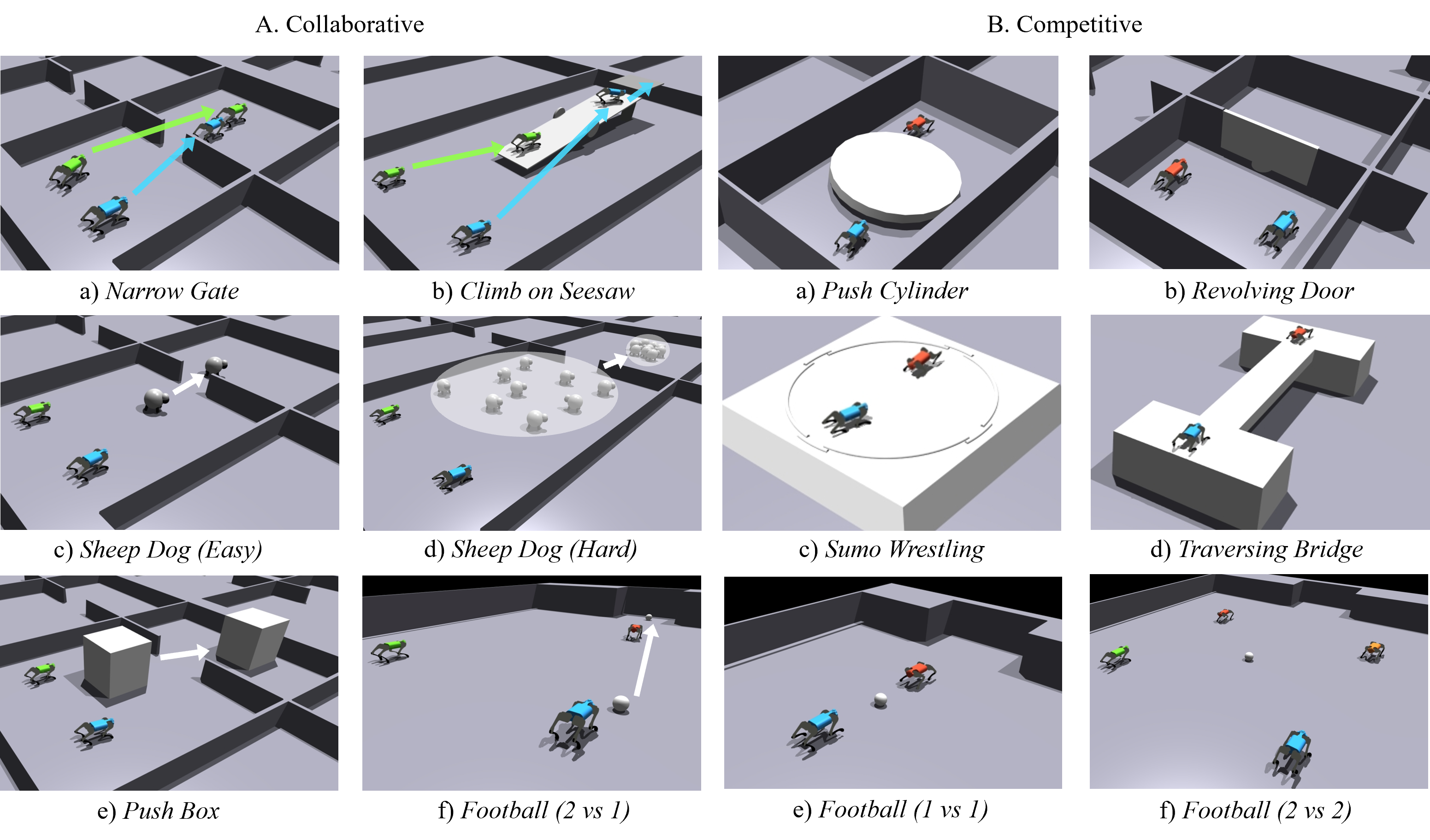}}
\caption{Demonstration of benchmarking tasks. \textcolor{blue}{Blue} and \textcolor{green}{green} robots are assigned to accomplish collaborative tasks, while \red{red} and \textcolor{orange}{orange} robots will play against them. Arrows of different colors illustrate the intended movements of agents and objects in each task. Generally, tasks demonstrated lower are harder due to the rising demands of advanced locomotion control and environment awareness.}
\label{demonstration}
\end{figure*}

\paragraph{Alternative hierarchical policy} Recognizing the challenges associated with learning a locomotion policy from scratch in a multi-agent task setting, our environment incorporates a robust walking policy as an optional foundational layer. This low-level policy responds to commands from the high-level policy. This structure allows for the simplification of some tasks as straightforward multi-agent tasks with complex movements, given that robot dynamics might not strictly follow the commands due to potential collisions, sliding, or command alterations.

\paragraph{Parallel Simulation} Leveraging the GPU-accelerated capabilities of Isaac Gym, our environment supports the execution of numerous simulations in parallel. This feature significantly enhances the sample efficiency of reinforcement learning, facilitating the achievement of high-performance policy within a constrained time.

\section{Benchamarking Tasks}

Based on the previously outlined environment, we have designed 12 tasks with varying degrees of complexity and distinct characteristics for benchmarking purposes. These tasks are evenly divided into collaborative and competitive categories, with their details illustrated in Figure \ref{demonstration}. The objective of collaborative tasks includes: 1) requiring agents to adopt asymmetric policies to achieve their goals, i.e. creating space to prevent potential collisions, and 2) necessitating cooperation among agents, i.e. passing and stopping a football. On the other hand, competitive tasks are designed to foster self-improvement among agents through self-play, as opposed to learning in a solitary agent setting, i.e. dribbling a football under the opposing defender's defense.

Actions are torques of the robot's actuated Degrees of Freedom (DoFs), or target positions that will be converted to torques through PD control. To accelerate the learning of high-level planning, we forgo the extensive training of a locomotion policy in favor of incorporating a pre-trained walk policy specifically designed for the Unitree Go1 robot ~\cite{margolis2023walk}. This advanced locomotion policy takes 18 dimensions of command and robot states, subsequently determining the target positions for each actuated DoF, which enables precise control over the Go1 robot's movement, ensuring it walks according to the specified gait and commands. For uniformity and simplicity in our benchmarking results, we will exclusively utilize the Unitree Go1 robot.

All collaborative benchmarking tasks are formulated as decentralized partially observable Markov Decision Processes (Dec-POMDPs) with selected observation and shared rewards, although the simulator itself is fully observable. We provide all available observations within the simulator to facilitate privileged observation in actor-critic algorithms or policy distillation. Proprioception is provided to the aforementioned low-level locomotion policy and thereby is omitted from the observations delivered to the high-level policy within the benchmarking tasks. This omission aims to evaluate the effectiveness of a hierarchical policy framework, where the high-level policy operates independently from low-level locomotion control.

\subsection{Collaborative Tasks}

\paragraph{Narrow Gate} A narrow gate that permits only one robot to go through splits a room. Positioned on the same side, two robots are instructed to walk through this gate sequentially, avoiding any collisions. This task necessitates an understanding of potential collisions and the management of routine to avoid collisions, requiring different agents to learn asymmetric policies of the timing and order for optimal rewards.

\paragraph{Climb on Seesaw} Positioned next to a suspended platform, a flat seesaw's design allows one end to be elevated to the platform's height, allowing the robot to climb onto the platform. Due to the mechanism of the seesaw, successful ascension requires coordination wherein one robot occupies the farther end of the seesaw, maintaining its stability, while the second robot ascends along the seesaw. The absence of this synchrony results in the seesaw collapsing.

\paragraph{Sheep Dog (Easy)} In \emph{Sheep Dog} tasks, we introduce sheep as non-player characters (NPCs) that exhibit autonomous movement based on the following principles: 1) sheep always maintain a distance from dogs, which are the robots controlled by agents, and will accelerate to keep away from the dogs, 2) sheep always tends to get close to the central of the herd, and 3) sheep always move with randomness in velocity. Similar to \emph{Narrow Gate}, \emph{Sheep Dog-Easy} introduces a sheep being placed on the same side as the dogs. The objective of this task is to guide the sheep through the gate by utilizing the movement features of sheep.

\paragraph{Sheep Dog (Hard)} Building on \emph{Sheep Dog (Easy)}, \emph{Sheep Dog-Hard} escalates the challenge by requiring the dogs to herd nine sheep, significantly increasing the complexity of the sheep's movements. This complexity challenges agents' perception of the position and movement of sheep and their corresponding ability to really "understand" the situation. It necessitates the agents follow the sheep's behavior patterns rather than relying on the memorization of the specific trajectories of sheep and dogs.

\paragraph{Push Box} Similar to the design of the \emph{Sheep Dog} tasks, \emph{Push Box} utilizes the terrain established in \emph{Narrow Gate}, with the sheep replaced by a heavy box that requires the collaborative effort of two robots to push. This task challenges to navigate and push the box through the gate. Due to the weight of the box, a dedicated locomotion policy for the pushing task will simplify the tasks. Furthermore, the task asks for cooperative behavior from both robots to maintain the box's trajectory and prevent its rotation.

\paragraph{Football (2 vs 1)} We create a football field enclosed by walls which replace sidelines and goals for \emph{Football} tasks. A defender, which is a robot serving as an NPC in this task, is programmed to position itself at the midpoint between the ball and the goal. Two robots are placed on the opposite side against the defender with a football. The target of the two robots is to kick the football into the goal. With the defender maintaining a position between the goal and the ball, this task not only requires the two dogs to possess dribbling skill but also necessitates strategic passing to circumvent the defender's blockades and successfully score, demanding both refined locomotion policies and effective cooperative policy formulation.

\subsection{Competitive Tasks}

\begin{figure*}[tbp]
\centerline{\includegraphics[width=\textwidth]{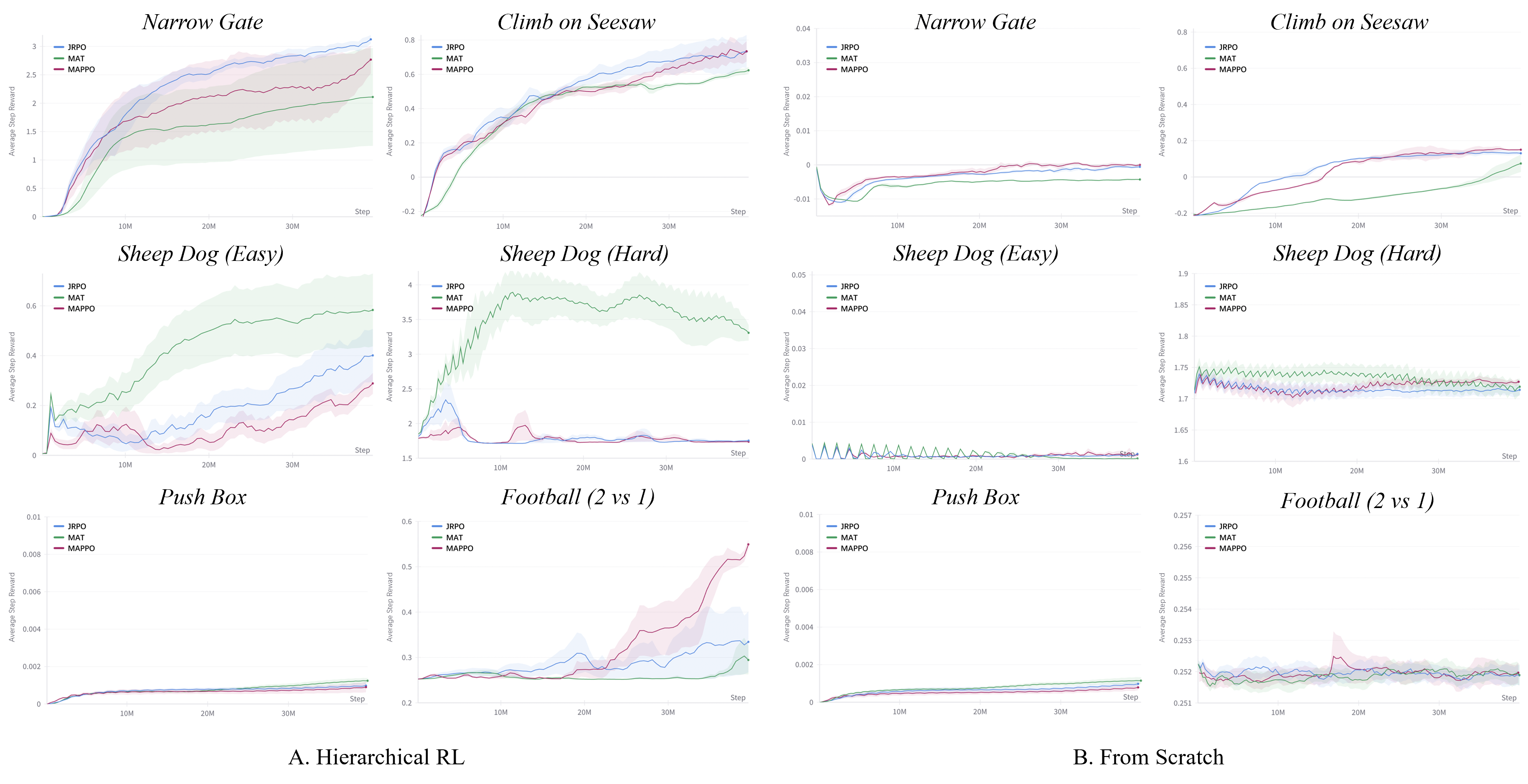}}
\caption{Learning curves of 6 tasks under 2 settings: results using pre-trained locomotion policy is illustrated left and results learning from scratch is illustrated right. Rewards of tasks related to sheep and boxes are jiggling on a small scale due to the reset of corresponding objects.}
\label{results}
\end{figure*}

\paragraph{Push Cylinder} A cylinder centrally placed in an arena can only move sideways. Two quadruped robots, positioned on opposite sides, compete by pushing the cylinder towards each other. The robot that pushes the cylinder towards the opponent's side earns a reward. This task challenges the robots to identify optimal directions and positions for pushing, leveraging the cylinder's circular cross-section for axial force.

\paragraph{Revolving Door} Centered on a wall, a revolving gate permits rotation and passage from both sides, albeit one at a time. Two robots placed on the same side compete to be the first to pass through the door by manipulating the door's rotation and the wall close to the door. This task demands that the robots understand the revolving door's mechanics and strategically counter their opponent's moves.

\paragraph{Sumo Wrestling} Two robots are placed in a sumo wrestling match on a featured platform. The objective for each robot is twofold: to destabilize and cause their opponent to fall or to push them out of the designated circle.  This task challenges the robots' ability to maintain balance when subjected to external forces, as well as their capability to tactically disrupt the opponent's movement without compromising their own stability.

\paragraph{Traverse Bridge} With the intention of crossing a narrow bridge to the other side, two robots are placed at opposite ends. The bridge's limited width permits only one robot to pass at a time. Thus, the primary objective for each robot is to push its opponent off the bridge to secure passage, asking the robots to make use of every inch of space in maintaining equilibrium while attempting to displace the opposing robot.

\paragraph{Football (1 vs 1)} On the football field introduced in \emph{Football (2 vs 1)}, we ask two robots to compete for control of the football and try to goal, requiring the robots with capabilities to dribble, tackle and shoot. Goals only happen when there is a huge strength gap between two robots.

\paragraph{Football (2 vs 2)} By increasing the number of robots on both sides, \emph{Football (1 vs 1)} becomes a simple football match. The target of each side is simply winning the match by a goal, utilizing the ability of dribbling, passing, tackling, shooting and cohesive team collaboration. There are many complex strategies in a 2 vs 2 football game, making this task a challenge for both locomotion control and strategies.

\section{Experiments}

With the help of the pre-trained locomotion policy, our environment serves as a fair comparative basis for different MARL algorithms, as we get rid of the possible variance coming from different locomotion policies and focus fully on the planning capability provided by MARL algorithms. In the following sections, we showcase the performance of different multi-agent reinforcement learning algorithms using the 6 collaborative tasks with and without the pre-trained locomotion policy.

\subsection{Simulation Performance}

We select 4 tasks with different objects to illustrate the efficient simulation capabilities of our environment, which are \emph{Narrow Gate}, \emph{Push Box}, \emph{Sheep Dog (Easy)}.

\begin{table}[htbp]
\caption{Simulation Performance of Our Environment in FPS}
\begin{center}
\begin{tabular}{cccc}
\hline
\multirow{2}{*}{\# Envs} & \emph{Narrow Gate}$^*$ & \emph{Climb on Seesaw}$^*$ & \emph{Sheep Dog (Easy)}$^*$ \\
 & 2 agents & 2 agents \& 1 object & 2 agents \& 1 NPC \\
\hline
500 & $18307\pm 154$ & $14083\pm 63$ & $16164\pm 68$ \\ 
1000 & $27720\pm 80$ & $22856\pm 80$ & $25129\pm 82$ \\
\hline
\multicolumn{3}{l}{$^*$with the pre-trained locomotion policy running.}
\end{tabular}
\label{FPS}
\end{center}
\end{table}

Thanks to Isaac Gym's high-performance parallel simulating capabilities, our environment can achieve $10000+$ frames per second with pre-trained locomotion policy running, which contains two MLPs with 4 layers and 3 layers. Means and standard deviations of FPS (frame per second) are shown in Table \ref{FPS}, and all the results are obtained on a server with NVIDIA RTX3090, CUDA 11.4, and Isaac Gym Preview 4. The high speed of simulation extremely accelerates the learning process in the wall-clock time, with all data processes accomplished on GPUs.

\subsection{Environment Setup}

Currently, we evaluate the performance of MAPPO, JRPO, and MAT implementation in OpenRL~\cite{huang2023openrl} on all collaborative tasks. Reward specifications are listed in Table \ref{Rewards}, while two settings of actions are 1) commands of horizontal velocity and rotation velocity of yaw or 2) target positions of actuated DoFs, aiming to simulate two reinforcement learning settings: 1) utilizing pre-trained locomotion policy and hierarchical structure to fully focus on the high-level collaboration, and 2) learning from scratch. Subtle rewards designed for learning a walking policy are excluded because such rewards require sampling of target velocity~\cite{margolis2023walk} that is impossible to achieve in a task-oriented environment. We train each algorithm in 500 parallel environments for 40 Million environment steps with 5 different random seeds.

\subsection{Quantitative Results}

The performance of each algorithm on different tasks is shown in Figure \ref{results}. By comparing the results under two settings, it's undeniable that hierarchical structure greatly simplifies tasks that can be solved by utilizing pre-trained locomotion policy through high-level commands, i.e. \emph{Narrow Gate} and \emph{Climb on Seesaw} which only requires different agents to simply walk around, but it's impossible to solve tasks that demand more specialized locomotion policies like \emph{Push Box} and \emph{Football-Defender} that, intuitively, requires a policy for pushing instead of walking and policies for dribbling, passing and stopping footballs to successfully solve the tasks. And the poor performance of RL from scratch demonstrates that end-to-end learning in multi-robot tasks is impossible. Because of the intricate NPC introduced in \emph{Sheep Dog} tasks, MAT achieves notably better performance compared to JRPO and PPO, illustrating that environment awareness takes an important role in realistic multi-agent settings.

According to the results, we can divide the 6 collaborative tasks into 3 categories with rising complexity: 1) tasks that require the agents to perceive other agents' positions and collaborate according to the positions, including \emph{Narrow Gate} and \emph{Climb on Seesaw}; 2) tasks that require the agents to perceive both other agents' positions and have a general knowledge of how the environment will change according to all agents' involvement, i.e. \emph{Sheep Dog-Easy} and \emph{Sheep Dog-Hard}; and 3) tasks that demand specialized locomotion policy as well as collaboration capabilities required by previous 2 categories, including \emph{Push Box} and \emph{Football-Defender}. The different complexity between categories implies that a more sophisticated method of hierarchical RL other than simply outputting state-unrelated commands is necessary.

\section{Conclusion and Future Work}

In this paper, we introduced the Multi-agent Quadruped Environment, our new reinforcement learning environment that supports the convenient design of multi-agent tasks for quadruped robots, along with twelve tasks predefined, covering cooperative, adversarial, low-level control, and high-level planning learning. Leveraging the powerful parallel computing capabilities of IsaacGym, our environment is capable of simulating hundreds of environments in parallel, significantly accelerating the process of reinforcement learning. Within our environment, we explored the feasibility of hierarchical RL in the control of quadruped robots and easily completed some cooperative tasks with the help of a low-level locomotion policy. This demonstrates the viability of hierarchical RL in the field of quadruped robots and its relative speed advantage over end-to-end learning. However, the naive hierarchical methods we employed were insufficient for solving complex tasks, prompting us to explore and discuss some viable future work here.

\emph{a) Environment awareness} plays a critical role in tasks featuring with complex dynamics. However, the lack of observations in the real world as well as the difficulty in understanding dynamics, highlights the importance of enhancing environment perception and prediction capabilities in multi-agent tasks, marking it as a promising area for future research.

\emph{b) Varied low-level policies} will greatly simplify the process of solving any task through a hierarchical structure. We aspire to see an increasing number of low-level policies proven to be effective in the hierarchical structure as well as novel tasks demanding higher agility of quadrupeds.

\section*{Appendix}

Rewards for collaborative tasks are listed in Table \ref{Rewards}, where quadratic terms will be labeled by $^2$ and exponential terms will be labeled by $\exp$. 

\begin{table}[htbp]
\caption{Reward Specification}
\begin{center}
\begin{tabular}{ccc}
\hline
Task & Description of Reward & Scale  \\
\hline
\multirow{3}{*}{\emph{Narrow Gate}} & agent walked through the gate & 5. \\
 & decreased distance to the gate & 1. \\
 & distance between agents (clipped)$^2$ & 0.025 \\
 & collision penalty & -2 \\
\hline
\multirow{7}{*}{\emph{Climb on Seesaw}} & agent reached the destination & 10. \\
& height of agent & 1. \\
& agent's movement along the seesaw & 5 \\
& agent's distance from the seesaw$^2$ & -0.5 \\
& collision penalty & -2. \\
& falling penalty & -2. \\
& distance between agents (clipped)$^2$ & 0.25 \\
\hline
\multirow{2}{*}{\emph{Sheep Dog (Easy)}} & sheep moved through the gate & 1. \\
& decreased sheep's distance to the gate & 2. \\
\hline
\multirow{2}{*}{\emph{Sheep Dog (Hard)}} & sheep moved through the gate & 1. \\
& $\exp$(-sheep's distance to the gate) & 1. \\
\hline
\multirow{2}{*}{\emph{Push Box}} & box moved through the gate & 1. \\
& decreased box's distance to the gate & 10. \\
\hline
\multirow{2}{*}{\emph{Football (2 vs 1)}} & goal & 10. \\
& $\exp$(distance between ball and the goal) & 3. \\
\hline

\end{tabular}
\label{Rewards}
\end{center}
\end{table}

\begin{figure}[htbp]
\centerline{\includegraphics[width=0.48\textwidth]{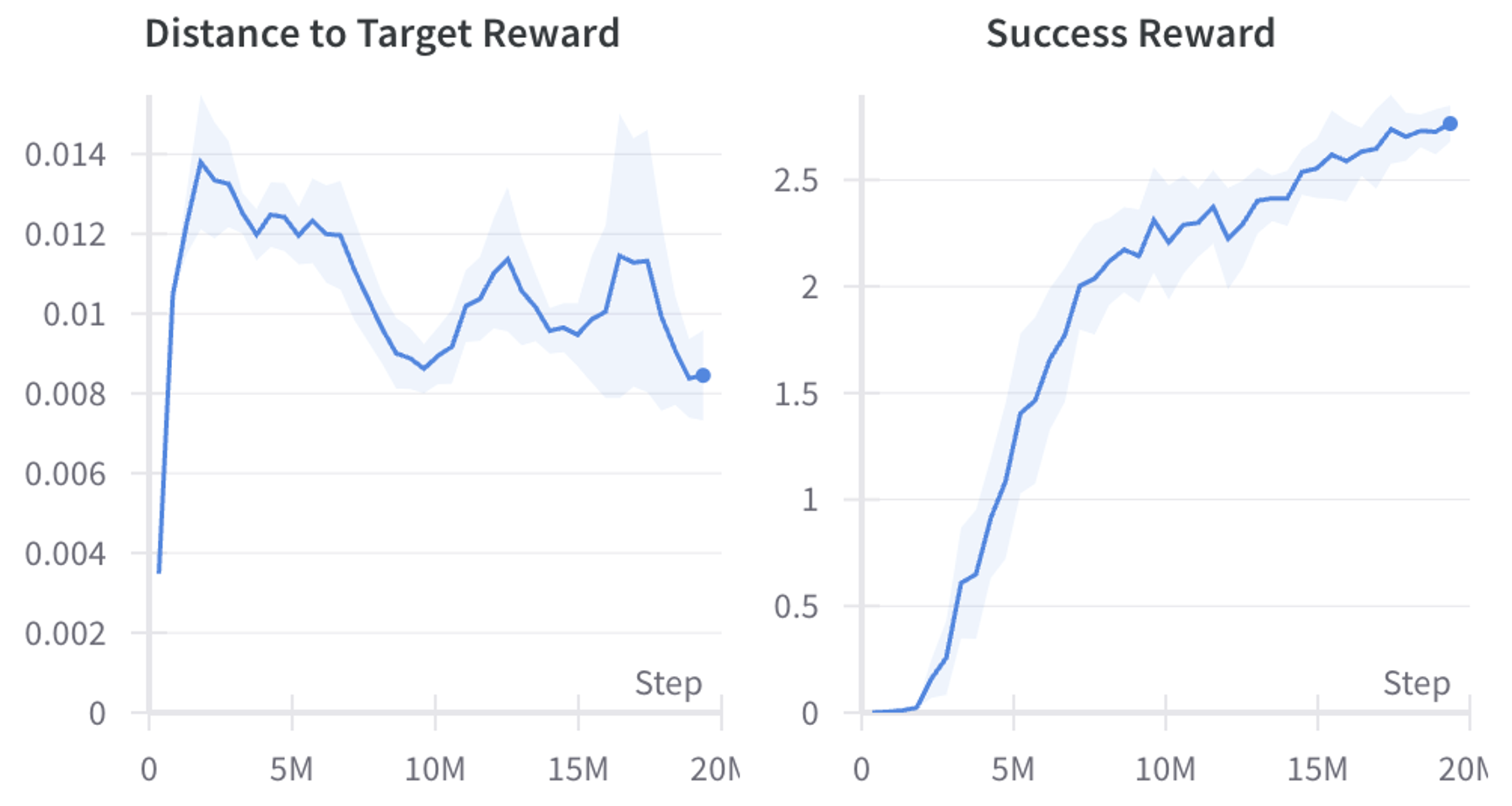}}
\caption{Learning curves of two selected rewards in PPO learning of \emph{Narrow Gate} within first 20M environment steps (average of 5 seeds).}
\label{appendix}
\end{figure}

There are two kinds of rewards: 1) rewards calculated based on change of states, i.e. Distance to Target Reward based on the decreased distance between agents and a specified target point behind the gate in \emph{Narrow Gate}, are designed to guide the agent at the beginning stage of learning as they are easy to acquire, and 2) rewards calculated based on states, i.e. Success Reward in \emph{Narrow Gate} provided when the agent has crossed the gate, are always closely related to success states of the tasks and result in much higher returns as they are provided repeatedly. Intuitively, the agent will first follow the change-of-state-based rewards, which may not provide an optimal guide to the target state, to discover the state space including success states, and then the state-based rewards illustrating the success states will dominate the learning. Figure \ref{appendix} shows the effectiveness of the first kind of reward at the beginning stage and then the domination of the second kind of reward.

\newpage

\bibliography{ieee.bib}
\bibliographystyle{IEEEtran}

\ifCLASSOPTIONcaptionsoff
  \newpage
\fi

\end{document}